\documentclass[11pt,a4paper]{article}
\usepackage[hyperref]{acl2020}
\usepackage{times}
\usepackage{latexsym}

\usepackage{graphicx}
\usepackage{amsmath}
\usepackage{tikz}
\usepackage{tikz-dependency}
\usepackage{microtype}

\aclfinalcopy 

\title{Improving cross-lingual model transfer by chunking}

\author{Ayan Das \\
  Dept. of Computer Science and Engg. \\
  IIT Kharagpur, India \\
  \texttt{ayan.das@cse.iitkgp.ac.in} \\\And
  Sudeshna Sarkar \\
  Dept. of Computer Science and Engg. \\
  IIT Kharagpur, India  \\
  \texttt{sudeshna@cse.iitkgp.ac.in} \\}

\date{}

\begin{document}
\maketitle
\begin{abstract}
We present a shallow parser guided cross-lingual model transfer approach in order to address the syntactic differences between source and target languages more effectively. In this work, we assume the chunks or phrases in a sentence as transfer units in order to address the syntactic differences between the source and target languages arising due to the differences in ordering of words in the phrases and the ordering of phrases in a sentence separately. 

\end{abstract}

\section{Introduction}
Model transfer approaches for cross-lingual dependency parsing involve training a parser model using a treebank of a language (source language) and using it to parse sentences of another language (target language). This technique may be used to develop parsers for languages for which no treebank is available.

The performance of cross-lingual parser models often tend to suffer due to the syntactic difference between the source and the target languages~\cite{zeman2008,sogaard2011,naseem2012}. Thus, a major challenge in transfer parsing is to bridge the gap between the source and target language. For example, the adjectives appear before the corresponding nouns in English and Hindi while in Spanish and Arabic the adjectives appear after the nouns. Several approaches have been proposed to address the syntactic differences. These include training a parser model using the a selected subset of source language parse trees that are syntactically close to the target language~\cite{sogaard2011, wang2016}, transformation of source language treebank to match the syntax of the target language~\cite{aufrant2016,das2019,wang2018b}, target-language independent perturbation~\cite{das2019a}, training a word order-insensitive parser model~\cite{ahmad2019} or imposing target-language syntax based constraints while running MST on the edge-score matrix of a graph-based parser to obtain the target language parse tree~\cite{meng2019}.

The syntax of a language may be classified into two categories. Firstly, the syntax of the words within the chunks or phrases (\textit{intra-chunk} syntax) and secondly, the orientation of the chunks in a sentence (\textit{inter-chunk} syntax). Consider the following English sentence.\\
\noindent EN: \textit{(The US) (lost) (yet another helicopter) (to hostile fire)}\\
The word groups enclosed by brackets indicate separate chunks or phrases and \textit{US, helicopter} and \textit{fire} are the head words of the chunks \textit{the US}, \textit{yet another helicopter} and \textit{to hostile fire} respectively. In this example, the \textit{intra-chunk} phrase syntax corresponds to the relative ordering of the determiners, adpositions, adjevtival modifiers, auxiliaries etc. with respect to the head words in a phrase whereas the the \textit{inter-chunk} syntax corresponds to the relative ordering of the chunks in the sentence.

Given a source-target language pair, the syntactic differences may be in the ordering of the words within a phrase, or, in the orientation of the phrases in sentence, or both. 
For example, the adpositions appear before the corresponding nouns in English while they appear after the corresponding nouns in Hindi. These differences are local to the phrases.  Similarly, languages also differ in the orientation of the phrases in the sentences. For example, the English, French, Spanish etc. follows SVO ordering, Japanese, Urdu and Hindi typically follow  SOV ordering, while, Arabic and Irish predominantly follows VSO ordering.

Consider the English sentence and its Hindi translation. \\
\noindent EN: \textit{``(He) (teaches) (the children)"}\\
  HI: \textit{``(va) (bachchOM kO) (paDhAtA hEi)''}\\
  EN-gloss: \textit{``(He) (children to) (teaches is)''}\\

Here, the phrase ``\textit{the children}" maps to the Hindi phrase \textit{bachchOM kO (children to)} and the English phrase \textit{``teaches"} maps to the Hindi phrase \textit{``paDhAtA hEi" (teaches is)}. We observe that the phrases have the following differences. The definite article is absent in Hindi. In Hindi, the postposition \textit{ko} is associated with the word \textit{bachchOM (children)} while no adposition is associated with the word \textit{children} in the corresponding English phrase. In the Hindi verb phrase, \textit{paDhAtA (teaches)} is followed by the copula verb \textit{hEi (is)}. Furthermore, the English sentence follows a SVO ordering of phrases while the Hindi sentence is verb-final.

Both the intra-phrase and inter-phrase differences affect the performance of the transfer parsers. Thus in order to simplify the transfer process we address these two differences separately. We propose to carry out a chunk information guided cross-lingual model transfer for dependency parsing where we treat the chunks as transfer units instead of the words. To this end, we train a source language parser model using the chunks as units. Given a target language sentence, the source language parser model is used to parse the target language chunks followed by the expansion of the target language chunks to obtain the complete trees. 

We propose to use chunk information in transfer parsing because a chunker (shallow parser) may be trained using lesser amount of data as compared to a full syntactic parser. Annotating data for a chunker is also much simpler as compared to that of a parser. The chunkers may also be rule-based whose development do not require any data.

\section{Related work}
Chunking (shallow parsing) has been used successfully to develop good quality parsers in Hindi language~\cite{bharati2009b,chatterji2012}. \newcite{bharati2009b} have proposed a two-stage 
constraint-based approach where they first tried to extract the intra-chunk dependencies and resolve the inter-chunk dependencies in the second stage. They have also shown effect of hard and soft constraints to build efficient Hindi parser that outperforms data-driven parsers.

\newcite{ambati2010b} used disjoint sets dependency relation and performed the intra-chunk 
parsing and inter-chunk parsing separately. ~\newcite{chatterji2012} proposed a three stage approach where a rule-based inter-chunk parsing followed a data-driven inter-chunk parsing.
 
A project for building multi-representational and multi-layered treebanks for Hindi and Urdu~\cite{bhatt2009a} \footnote{http://verbs.colorado.edu/hindiurdu/index.html} was carried out as a joint effort by IIIT Hyderabad, University of Colorado and University of Washington.
Besides the syntactic version of the treebank being developed by IIIT Hyderabad~\cite{ambati2011}, University of Colorado has built the Hindi-Urdu proposition bank~\cite{vaidya2014} and a phrase-structure form of the treebank~\cite{bhatt2012} is being
developed at University of Washington. A part of the Hindi dependency treebank\footnote{http://ltrc.iiit.ac.in/treebank\_H2014/} has been released in which the inter-chunk dependency relations (dependency links between chunk heads) have been manually tagged and the chunks were expanded automatically using an arc-eager algorithm. Some of the major works on parsing in Bengali language appeared in ICON 2009 (http://www.icon2009.in/). 
\newcite{ghosh2009} used a CRF based hybrid method, ~\newcite{chatterji2009} used variations of the transition based dependency parsing. \newcite{mannem2009} came up with a bi-directional  incremental  parsing  and  perceptron  learning approach and  \newcite{de2009} used a constraint-based method. \newcite{garain2012} compares performance of a grammar driven parser
 and a modified MALT parser.

\section{Chunking}
\label{subsec:chunktrans-chunking}
Chunking involves identification of different phrases in a sentence and identification of a chunk-head or main word in a given chunk. A chunker may be rule-based or data-driven. In a rule-based chunker a set of pre-defined rules are used to identify the chunks and the corresponding heads. On the other hand in a data-driven chunker, the task of chunking is usually posed a sequence labelling task and a machine learning-based algorithm is trained for chunk identification.

On the other hand, rule-based approaches are usually used for chunk head identification.
\begin{figure*}[htbp]
    \centering
    \includegraphics[scale=0.65]{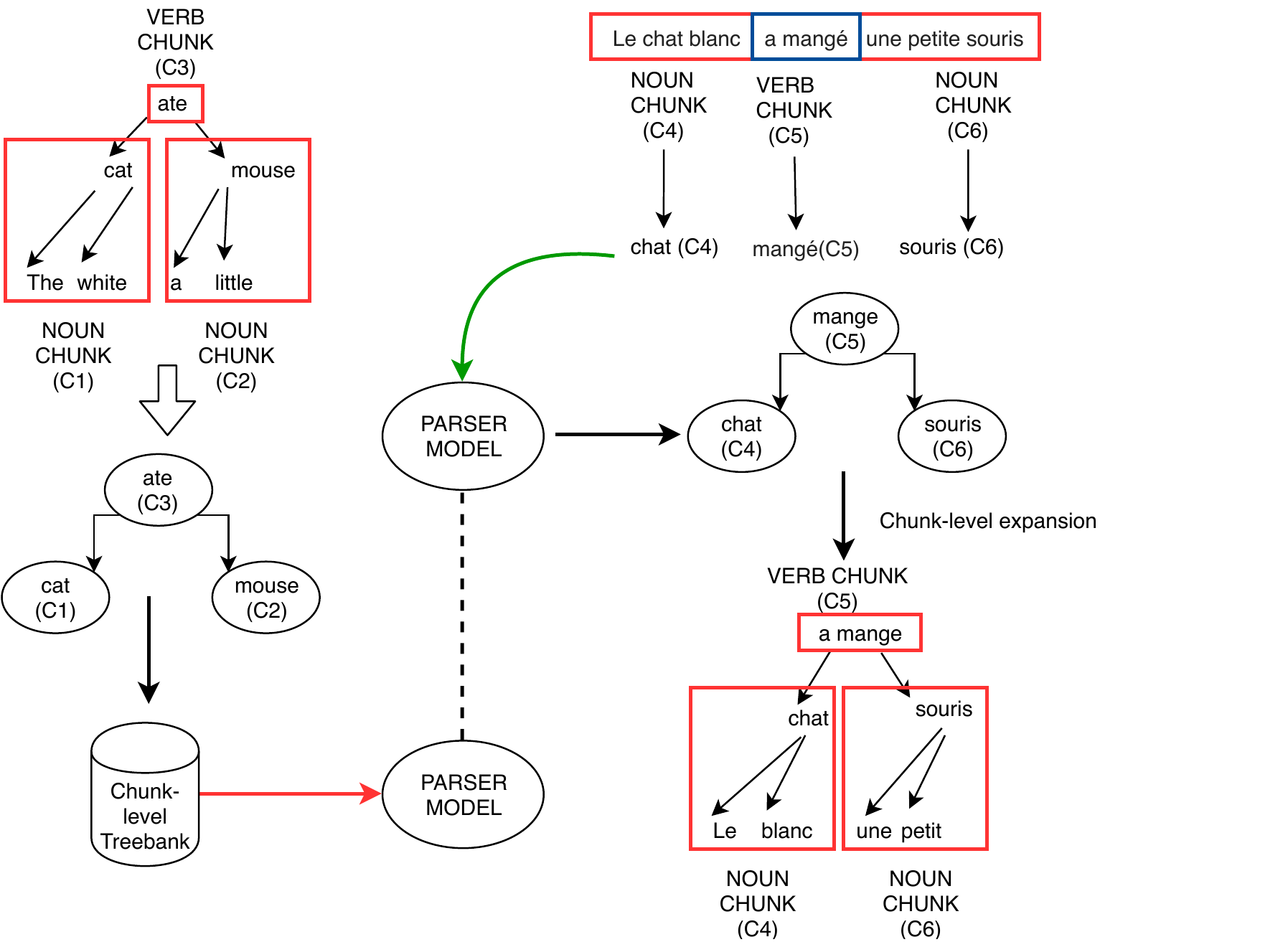}
    \caption{\label{fig:chunktrans-chunktransfer1} Chunk-level model transfer for dependency parsing}
\end{figure*}

\subsection{Chunk identification}
In this work, we address the problem of chunk identification as a sequence labelling task where we label each chunk using the \textit{BI} labelling e.g. in the above example the chunk sequence is as follows;\\
\noindent EN: (The US) (lost) (yet another helicopter) (to hostile fire)\\
\noindent Labels: B-NP I-NP B-VP B-NP I-NP I-NP B-NP I-NP I-NP\\
where \textit{B-*} indicates the beginning of a chunk an \textit{I-*} indicates inside the chunk starting at the \textit{B-*} immediately preceding it. In this work we have identified the following chunk types: noun phrase (NP), verb phrase (VP), adjectival phrase  (JJP), adverbial phrase (RBP), coordinating conjunctive phrase (CCP) and the remaining as BLK.

The chunk-type is determined based on the PoS tag of the chunk-head word e.g. if the chunk head word is a noun, pronoun or proper noun then it is assigned the \textit{NP} chunk type. The beginning of a chunk is not necessarily the chunk head. In Table~\ref{tab:chunktrans-chunkexample} we present the chunk annotation of an example sentence. The subscripts in the last column indicates the chunk number in the sentence.

\begin{table}[htbp]
    \centering
    \begin{tabular}{c|c|c|c}
    \hline
        Word &  PoS & Chunk type & Head type\\
        \hline
         The & DET  & B-NP  & child$_1$  \\
         white & ADJ & I-NP  & child$_1$  \\
         cat & NOUN &  I-NP & head$_1$  \\
         \hline
         ate & VERB & B-VP  &  head$_2$ \\
         \hline
         a & DET &  B-NP &  child$_3$ \\ 
         little & ADJ & I-NP  &  child$_3$ \\ 
         mouse & NOUN & I-NP  &  head$_3$ \\ 
        \hline
    \end{tabular}
    \caption{\label{tab:chunktrans-chunkexample} Chunking example.}
\end{table}

\subsubsection{Chunker model} We used a CRF-BiLSTM based neural model to train the chunker. The 2-layer bi-directional LSTM takes the embeddings of the PoS tags of the words in a sentence as input and encodes them in the internal states. We used the hidden states corresponding to the final layers of the forward and backward LSTMs as the distributed representation of the corresponding words. These word representations were used as input to a CRF for chunk-label prediction.

\subsection{Chunk head identification}
We used a rule-based approach for predicting the chunk-head in a given chunk. Based on the category of a given chunk we designed a set of rules for predicting the most probable head. The set of rules varies slightly across languages.

\section{Chunking based cross-lingual model transfer}
In this section, we present our approach for shallow parser-guided cross-lingual transfer parsing where the transfer is carried out at the chunk-level instead of the word-level. In Figure~\ref{fig:chunktrans-chunktransfer1} we show a schematic diagram of the steps of our chunk-level model transfer.

This method requires training a source language parser model using chunks as unit and a shallow parser in the target language.

Training a chunk-level source language parser model involves derivation of the chunk-level parse trees and training the parser model using the chunk-level parse trees. In case of source language, the chunk-level parse trees are derived using the chunk annotation of the training data. This is done by collapsing the sub-trees corresponding to each chunk and replacing them by a chunk representation. The chunk-level parse tree so obtained is used to train the parser model.

Given a target language sentence, the shallow parser is used to identify the chunks in the sentence. In case of the target language, the chunk representations are obtained by simply replacing the words in a chunk by a representation. The sequence of chunk representations so obtained are parsed using the chunk-level source language parser model. Finally, the target language chunks are expanded to obtain the full target language parse tree.

We elaborate the steps for training a chunk-level transfer and parsing target language sentences using the model in Section~\ref{subsec:chunktrans-trainingchunkparser} and \ref{subsec:chunktrans-parsingchunkseq} below.

\subsection{Training a chunk-level parser model}
\label{subsec:chunktrans-trainingchunkparser}
The steps for training a chunk-level transfer parser are as follows.
\subsubsection{Obtaining the chunk-level source language treebank} The chunk-level source language parse trees are derived from the parse trees in source language treebank by collapsing the chunks and replacing the chunks by their representations. Here we represent a chunk by its chunk head. In the example, the English sentence $(The\ white\ cat)_{NP} (\ ate)_{VP}\ (a\ little\ mouse)_{NP}$'', the chunks \textit{The white cat}, \textit{ate} and \textit{a little mouse} are collapsed and represented by their chunk heads \textit{cat}, \textit{ate} and \textit{mouse} respectively. In the parse tree, the relations corresponding intra-chunk words are also removed. The final tree consists of the chunk representations and the relations among them corresponds to the relations among the chunk heads as shown in the diagram.

\subsubsection{Training the chunk-level parser model} The chunk-level parse trees derived from the source language treebank in the above step are then used to train the parser model.

\subsection{Chunk-level parsing followed by chunk expansion}
\label{subsec:chunktrans-parsingchunkseq}
The steps for generating the parse tree for a given target language sentence are as follows.

\subsubsection{Chunking a target language sentence} A target language chunker is used to identify the chunks in a target language sentence and the heads of each chunk is identified  using a rule-based technique as discussed in Section~\ref{subsec:chunktrans-chunking}. Assuming French to be the target language, let us consider the sentence in the the example.\\
\noindent FR: \textit{``Le chat blanc a mange une petit souris''}\\
A chunker is used to identify identify the chunks as follows;
``(Le\ chat\ $\text{blanc})_{NP}$\ (a\ $\text{mange})_{VP}$\ (une\ petit\ $\text{souris})_{NP}$''. 

The rule based chunk head identifier is then used to identify the chunk heads. The heads of the chunks \textit{Le chat blanc}, \textit{a mange}, \textit{une petit souris} are \textit{chat}, \textit{mange} and \textit{souris} respectively.

\subsubsection{Parsing a target language chunk sequence}
The sequence of the target language chunk-head sequence obtained above is parsed using the parser model trained using the source language chunk-head parse trees to obtain the chunk-head parse tree as shown in the diagram.

\subsubsection{Chunk expansion}
The chunk-head parse tree so obtained is then expanded to obtained the parse tree of the target language sentence. To this end, we expand each chunk in the chunk-head parse tree by attaching the non-chunk-head word in the chunk to its corresponding chunk-head by a modifier-head relation without any change in relative ordering of words in the sentence. In this relation, the chunk-head is the head and the non-chunk-head word is the modifier.

As shown in the above example, the chunk represented by \textit{chat} is expanded by attaching the words \textit{Le} and \textit{blanc} to the chunk-head \textit{chat} to obtain the parse of the chunk.

As of now, we do not associate any dependency relation to the intra-chunk relations.

\section{Data and parser model}
\subsection{Data}
We carried out our experiments using English (\textit{en}) and Hindi (\textit{hi}) as source languages, and, English, French (\textit{fr}), German (\textit{de}), Indonesian (\textit{id}), Hebrew (\textit{he}), Arabic (\textit{ar}), Korean (\textit{ko}) and Hindi (\textit{hi}) as target languages. We used the UD v2.0 treebanks for our experiments.

\subsubsection{Data for training chunkers}
We trained our chunker using the gold annotations obtained from the UD 2.0 treebanks of the languages.

We classified the UD dependency relations into two groups; \textit{intra-chunk} and \textit{inter-chunk}. Our set of intra-chunk dependency relations comprises of the \textit{aux, appos, nummod, det, case, fixed, flat, compound, amod, advmod} and {goeswith} relations. The words related by the other dependency relations such as \textit{nsubj, obj, iobj, root, obl, comp, cc, conj} etc. were considered to be the chunk-heads and their relations with their parents were considered inter-chunk relations. In case of the \textit{amod} and \textit{advmod} relations, we selectively considered the dependents as intra-chunk. In case of \textit{amod}, dependents whose parents are \textit{nouns, adjectives} or \textit{adverbs} were considered as intra-chunks and in case of \textit{advmod}, the dependents with \textit{verbs, adverbs} and \textit{adjectives} as dependents were considered as intra-chunks. In a dependency parse tree, a chunk-head word along with all its dependents related to it by intra-chunk relations were considered to be a chunk.

\begin{table*}[htbp]
    \centering
    \begin{tabular}{c|c|c|c}
    \hline
    \shortstack{Training \\set size}&\shortstack{Avg. chunking \\acc.(\%)}& \shortstack{Avg. acc.\\ (En as src.)} & \shortstack{Avg. acc.\\ (Hi as src.)}\\
    \hline
    20      &  67.3&  46.8  & 37.1  \\
    50      &  75.3&  48.9  & 39.9  \\
    100     &  82.7&  54.4  & 42.8  \\
    200     &  86.1&  55.4  & 44.0  \\
    300     &  86.7&  55.9  & 44.7  \\
    500     &  88.1&  56.8  & 45.6  \\
    1000    &  88.6&  56.8  & 45.2  \\
    1500    &  89.3&  57.0  & 45.1  \\
    \hline
    Gold chunk &  100  & 65.9  & 55.2\\
    \hline
    Full tree& \_ & 48.0 & 41.5  \\
    \hline
    \end{tabular}
    \caption{\label{tab:chunktrans-chunkperfwithsize} Variation of average chunker accuracy and UAS of chunk-level transfer over 6 languages with training set size with English and Hindi as source languages}
\end{table*}


\subsubsection{Parser data}
The chunk-level parse trees were obtained by removing the sub-trees corresponding to the nodes having \textit{intra-chunk} relations with their parents. The removal of the phrase sub-trees left us with the skeleton trees in which all the words are chunk-heads and are related to their parents via inter-chunk relations. Thus in each chunk-level tree, each chunk is represented by their chunk head. We trained the chunk-level parser model using these chunk-level trees derived from the training partition of the treebank of source languages.

\subsection{Parser model}
For our experiments, we trained a transition-based encoder-decoder parser model that use a bi-directional LSTM as encoder and a attention-based decoder using stack-pointers~\cite{ma2018}.

\section{Experiments and results}
In this section, we discuss in details the experiments and results.

\subsection{Chunk labelling and chunk head identification}

\begin{table}[htbp]
    \centering
    \begin{tabular}{c|c}
    \hline
    Language & \shortstack{Chunk head\\ identification\\ accuracy}\\
    \hline
    en     &  97.5 \\
    fr     &  99.2 \\
    de     &  98.6 \\
    he     &  99.1 \\
    id     &  95.2 \\
    ar     &  99.1 \\
    ko     &  99.5 \\
    hi     &  98.7 \\
    ja     & 97.9  \\
    \hline
    Avg.   & 98.3  \\
    \hline
    \end{tabular}
    \caption{\label{tab:chunktrans-chunkheadaccuracy} Chunk head identification accuracies for different languages}
\end{table}

We experimented with different sizes of dataset for training the chunkers. In the second column of Table~\ref{tab:chunktrans-chunkperfwithsize} we report the average performance of the chunkers over the 9 languages corresponding to the different sizes of training data. We observe that the accuracy increases with training set size and stabilizes beyond a training set of 500 sentences.


In Table~\ref{tab:chunktrans-chunkheadaccuracy} we present the chunk-head identification accuracy for the different languages. We observe that although we have used a very simple rule set for chunk head identification, we achieved significantly high accuracies in chunk head identification.

\subsection{Chunk-level parsing}
\subsubsection{Baseline} We compare the performance of the chunk-level transfer models with the performance of the corresponding word-level transfer parser models as baseline. For both word-level and chunk-level transfer parsing we adopted the delexicalized transfer parser models.

\subsubsection{Chunk-level transfer parser}
We experimented with both predicted and gold annotations of the test data.
\begin{itemize}
    \item For predicted chunk annotation, the chunker models trained on 500 sentences were used to automatically label the test data and the chunk-heads were identified using the rule-based method discussed above.
    \item For the gold annotation, we directly used the gold chunk annotation of the test data.
\end{itemize}

\paragraph{Evaluation metric: }We report the results of our experiments in terms of unlabeled attachment score (UAS) and labelled attachment score (LAS).

\subsubsection{English as source language}
Here we discuss the performance of the chunk-level transfer parser approach with English as source language.

In the third column of Table~\ref{tab:chunktrans-chunkperfwithsize} we present the variation of the average UAS over the 9 target languages with training set sizes of the chunkers used to predict the chunks. We observe the beyond a training set size of 50 the average performance of chunk-level transfer parser improves over the performance on the word-level transfer. We also observe that the performance stabilizes at about a chunker training size of 500 sentences. In the following discussion with English as source language we report the results corresponding to the chunkers trained with 500 instances. 

In Table~\ref{tab:chunktrans-englishsrcfull} we compare the performance of the our chunk-level cross-lingual transfer parser model with the baseline. Since, we did not assign any relation type to the intra-chunk head-dependent dependency relations we report the UAS scores only for the full trees. In this table, we report results corresponding to predicted chunks and gold chunks. We have primarily compared the baseline with the performance of transfer parsing with predicted chunks. The bold entries indicates the higher of the UAS values. We have reported the results with gold chunks for reference in order to show the improvement in case gold annotated chunk information is available. We underlined the entries corresponding to the transfer parsers with gold chunks  where it gives the highest UAS among the three results corresponding to a language.

\begin{table}[htbp]
    \centering
    \begin{tabular}{c|c|c||c}
    \hline
    Lang&\shortstack{UAS\\with\\full tree\\ transfer} &\shortstack{UAS\\with\\predicted\\chunks} & \shortstack{UAS\\with\\gold\\chunks} \\
    \hline
    en& \textbf{94.3} & 73.7  & 89.5 \\
    fr& \textbf{76.1} & 74.1  & 80.0 \\
    de& \textbf{62.8} & 60.1  & \underline{73.3} \\
    he& 52.6 & \textbf{56.5}  & \underline{65.9} \\
    id& 46.3 &\textbf{63.2}   & \underline{71.3} \\
    ar& 30.3 & \textbf{50.2}  & \underline{52.6} \\
    ko& 27.9 & \textbf{36.8}  & \underline{47.2}  \\
    hi& 26.5 & \textbf{51.1}  & \underline{57.9} \\
    ja& 15.1 & \textbf{45.2}  & \underline{55.5} \\
    \hline
    Avg& 48.0 & \textbf{56.8}  & \underline{65.9} \\
    \hline
    \end{tabular}
    \caption{\label{tab:chunktrans-englishsrcfull} Comparison of performance of chunk-level transfer parser with the baseline transfer model for English as source language. The results are based target language chunks predicted using chunker trained on 500 sentences.}
\end{table}

In Table~\ref{tab:chunktrans-englishsrcinter} we compare the performance on the inter-chunk relations only. In this case we report both the UAS and LAS.

We observe that on an average across the 9 target languages our two-stage chunk-level transfer parser performs better than the baseline. Furthermore, it performs better than the baseline in case of 5 out of the 9 target languages. We also observe that the performance of our approach improves with the syntactic distance between the source and the target languages. We also observe that the chunk-level transfer parser with gold chunk information performs better than the baseline in case of 8 out of 9 target languages in terms of UAS and 7 languages in terms of LAS.

\begin{table}[htbp]
    \centering
    \begin{tabular}{c|c|c|c|c||c|c}
    \hline
    Lang& \multicolumn{2}{c|}{\shortstack{Full tree\\transfer}}   &   \multicolumn{2}{c||}{\shortstack{Predicted\\ chunk\\transfer}}   &  \multicolumn{2}{c}{\shortstack{Gold \\chunk\\transfer}}  \\
    \hline
    &  U &  L &   U &  L   &  U &  L  \\
    \hline
    en&  \textbf{88.5} &  \textbf{78.2} &  74.8 & 68.7  & \underline{90.3} & \underline{85.4}\\
    fr&  \textbf{68.9} & \textbf{48.4}  &  66.1 & 51.5  &  \underline{74.0} & \underline{56.9} \\
    de&  \textbf{51.6} & \textbf{39.7}  &  50.8 & 36.4  &  \underline{63.3} & \underline{48.7} \\
    he&  43.0 & 29.4  &  \textbf{53.5} & \textbf{31.2}  &  \underline{60.4} & \underline{38.8} \\
    id& 29.4  & 35.5  & \textbf{65.6}  & \textbf{53.9} &  \underline{75.3} & \underline{64.4} \\
    ar& 22.9  & 13.8  & \textbf{44.7}  &\textbf{ 26.6}  &  \underline{47.1} & \underline{29.5} \\
    ko&  20.0 & 10.5  & \textbf{36.4}  & \textbf{21.2} &  \underline{38.2} & \underline{24.0} \\
    hi& 28.2  & 19.0  & \textbf{32.3}  &\textbf{19.2}  &  \underline{39.1} & \underline{25.5} \\
    ja& 13.7  & 7.9  &  \textbf{21.4} &  \textbf{9.3} &  \underline{22.8} &  \underline{10.6}\\
    \hline
    Avg & 40.7  & 31.4  & \textbf{49.5}  & \textbf{35.3}  &\underline{56.7}  & \underline{42.6} \\
    \hline
    \end{tabular}
    \caption{\label{tab:chunktrans-englishsrcinter} Comparison of performance of chunk-level transfer parser with the baseline transfer model on the inter-chunk relations only with English as source language. The chunks were predicted using chunker was trained on 500 sentences.}
\end{table}

\subsubsection{Hindi as source language}
We repeated our experiments with Hindi as source language and the same set of target language as above. In the fourth column of Table~\ref{tab:chunktrans-chunkperfwithsize} we present the variation of the average UAS over the 9 target languages with training set sizes of the chunkers used to predict the chunks. We observe the performance starts improving beyond a chunker training set size of 100 sentences. Furthermore the highest accuracy is achieved at 500 tree. Hence, in our following discussions we report the results corresponding to the chunkers trained with 500 instances. 

In Table~\ref{tab:chunktrans-hindisrcfull} we compare the performance of our chunk-level transfer parser with the baseline on full trees and in Table\ref{tab:chunktrans-hindisrcinter} we report the results corresponding to the inter-chunk relations.
\begin{table}[!ht]
    \centering
    \begin{tabular}{c|c|c||c}
    \hline
    Lang&\shortstack{UAS\\with\\full tree\\ transfer} &\shortstack{UAS\\with\\predicted\\chunks} & \shortstack{UAS\\with\\gold\\chunks} \\
    \hline
         en& 39.7 & \textbf{45.3}  & \underline{57.2} \\
         fr& 32.7 & \textbf{48.8}  & \underline{53.1} \\
         de& 46.8 &  \textbf{50.8} & \underline{62.8} \\
         he& 24.5 & \textbf{33.3}  & \underline{39.9} \\
         id& 19.7 &  \textbf{31.2} & \underline{49.2} \\
         ar& 8.6 &  \textbf{10.2} & \underline{12.1} \\
         ko& 43.9 & \textbf{48.4}  & \underline{68.4} \\
         hi& \textbf{96.1} & 81.9  & 84.5 \\
         ja& \textbf{61.3} & 61.0  & \underline{69.7} \\
         \hline
         Avg& 41.5 & \textbf{45.6}  & \underline{55.2} \\
         \hline
    \end{tabular}
    \caption{\label{tab:chunktrans-hindisrcfull} Comparison of performance of chunk-level transfer parser with the baseline transfer model with Hindi as source language. The results are based target language chunks predicted using chunker trained on 500 sentences.}
\end{table}

From Table~\ref{tab:chunktrans-hindisrcfull} observe that corresponding to the all the full trees the chunk-level transfer followed by chunk expansion with predicted chunk information performs better than direct transfer with Hindi as source language for 7 out of 9 languages and also in terms of average performance over all target languages.

\begin{table}[htbp]
    \centering
    \begin{tabular}{c|c|c|c|c||c|c}
    Lang& \multicolumn{2}{c|}{\shortstack{Full tree\\transfer}}   &   \multicolumn{2}{c||}{\shortstack{Predicted\\ chunk\\transfer}}   &  \multicolumn{2}{c}{\shortstack{Gold \\chunk\\transfer}}  \\
    \hline
    &  U &  L &   U &  L   &  U &  L  \\
    \hline
    en & 37.6 & 22.2 & \textbf{36.7} & \textbf{25.2}   & \underline{40.1}  & \underline{27.7}   \\
    fr & 23.8 & 13.2 & \textbf{24.9} & \textbf{15.1}   & \underline{27.1}  &  \underline{17.8}  \\
    de & \textbf{44.5} & \textbf{30.5} & 39.6 &28.7  & \underline{46.5}  & \underline{32.7}   \\
    he & \textbf{24.9} & 8.9 & 23.0 & \textbf{9.4}    & \underline{25.7} & \underline{10.6}  \\
    id & 17.3 & 11.5 & \textbf{29.6} & \textbf{20.8}   & \underline{36.8}  & \underline{24.4}   \\
    ar & 7.2 & 4.1 & \textbf{10.1}    & \textbf{5.9}    &  \underline{10.6} &  \underline{6.9}  \\
    ko & \textbf{50.5} & \textbf{33.7} & 48.4 & 28.5   & \underline{61.9}  &  \underline{34.8}  \\
    hi & \textbf{95.1} & \textbf{91.1} & 89.7 & 81.9   & 90.8  & 84.2   \\
    ja & \textbf{50.4} & \textbf{27.7} & 49.0 & 25.1   &  \underline{51.4} & 27.0   \\
    \hline
    Avg& \textbf{39.0} & \textbf{27.0} & 38.2 & 25.1   & \underline{43.4}  &  \underline{29.5}  \\
    \hline
    \end{tabular}
    \caption{\label{tab:chunktrans-hindisrcinter} Comparison of performance of chunk-level transfer parser with the baseline transfer model on the inter-chunk relations only with Hindi as source language. The chunks were predicted using chunker was trained on 500 sentences.}
\end{table}

From Table~\ref{tab:chunktrans-hindisrcinter} we observe that with Hindi as source language the average performance of the chunk-level transfer with predicted chunk information is slightly worse than that of the baseline in terms of average UAS and LAS. However, it outperforms the baseline in case of 5 out of the 9 target languages.

\section{Conclusion}
\label{sec:chunktrans-chunktrans-conclusion}
In this chapter, we present an approach of cross-lingual transfer parsing that helps to reduce the error due to the syntactic differences between the source and target languages by addressing the intra-phrase and inter-phrase syntactic differences separately when chunkers are available for the two languages.

\bibliographystyle{acl_natbib}
\bibliography{main}
\end{document}